\documentclass[10pt,twocolumn,letterpaper]{article}

\usepackage{cvpr}              

\usepackage{lineno}
\usepackage{multirow}

\usepackage{booktabs}
\usepackage{xcolor}

\usepackage{algorithm}
\usepackage{algorithmic}
\usepackage{amsmath}
\DeclareMathOperator*{\argmax}{arg\,max}

\usepackage{makecell}
\usepackage{caption}
\usepackage{xcolor}

\usepackage{newfloat}
\usepackage{adjustbox}
\usepackage{listings}
\usepackage{stfloats}
\usepackage{afterpage}
\usepackage{cuted}

\definecolor{cvprblue}{rgb}{0.21,0.49,0.74}
\usepackage[pagebackref,breaklinks,colorlinks,allcolors=cvprblue]{hyperref}

\def\paperID{19} 
\def\confName{CVPR}
\def\confYear{2025}

\title{Kubrick: Multimodal Agent Collaborations for Synthetic Video Generation}

\author{Liu He$^{1}$ \quad Yizhi Song$^{1}$ \quad  Hejun Huang$^{2}$ \quad Pinxin Liu$^{3}$ \quad Yunlong Tang$^{3}$ \quad Daniel Aliaga$^{1}$ \quad Xin Zhou$^{2}$ \\
{\normalsize $^1$ Purdue University} \quad
{\normalsize $^2$ Baidu USA} \quad
{\normalsize $^3$ University of Rochester} \quad
}
\begin{document}

\twocolumn[{
\renewcommand\twocolumn[1][]{#1}
\maketitle
\vspace{-25pt}
\begin{center}
    \vspace{-5pt}
\includegraphics[width=\textwidth]{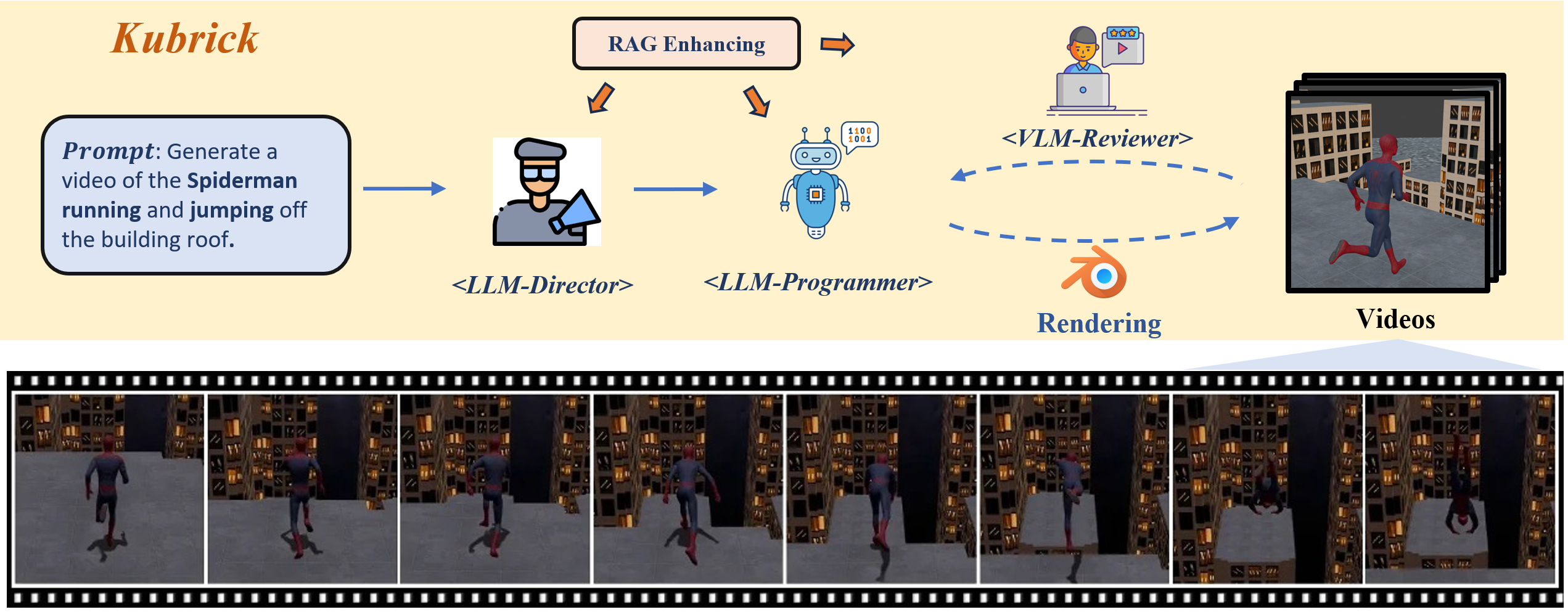}
    \vspace{-15pt}
    \captionof{figure}{\textbf{Kubrick Framework for Video Generation.} Kubrick consists of multiple agent collaborations, and iterative refinement of visual results. The \texttt{LLM-Director} decomposes the user-provided video description into functional sub-processes. Then, the \texttt{LLM-Programmer} compose Python scripts for Blender. Intermediate screenshots and video outputs will be reviewed by \texttt{VLM-Reviewer} for visual quality and prompt-following evaluation. The review will be fed back to \texttt{VLM-Reviewer} to iteratively improve the video. All three agents will be enhanced by RAG from online domain knowledge.}
    \label{fig:frameworks}
\end{center}
}]

\maketitle




\begin{abstract}
Text-to-video generation has been dominated by diffusion-based or autoregressive models. These novel models provide plausible versatility, but are criticized for improper physical motion, shading and illumination, camera motion, and temporal consistency. The film industry relies on manually-edited Computer-Generated Imagery (CGI) using 3D modeling software. Human-directed 3D synthetic videos address these shortcomings, but require tight collaboration between movie makers and 3D rendering experts. We introduce an automatic synthetic video generation pipeline based on Vision Large Language Model (VLM) agent collaborations. Given a language description of a video, multiple VLM agents direct various processes of the generation pipeline. They cooperate to create Blender scripts which render a video following the given description. Augmented with Blender-based movie making knowledge, the Director agent decomposes the text-based video description into sub-processes. For each sub-process, the Programmer agent produces Python-based Blender scripts based on function composing and API calling. The Reviewer agent, with knowledge of video reviewing, character motion coordinates, and intermediate screenshots, provides feedback to the Programmer agent. The Programmer agent iteratively improves scripts to yield the best video outcome. Our generated videos show better quality than commercial video generation models in five metrics on video quality and instruction-following performance. Our framework outperforms other approaches in a user study on quality, consistency, and rationality.

\end{abstract}

\section{Introduction}
\label{sec:intro}

Providing AI-assisted video generation is useful to facilitate drafting video sequences and synthetic animations. The video generation process should support high-level storyboard control of the video and handle the details of providing physically-based motions, proper shading and illumination, smooth camera motion, and temporal consistency. Thus, unlike the traditional tedious process of hand-crafted video editing limited to experts, a broader audience of users can focus on the storyboard and benefit from the AI-assisted methodology to produce high-quality videos of varying length and for a variety of content~\cite{huang2024scaling, song2024tri, liu2025gesturelsm}.

Prior methods can be divided into end-to-end video generation approaches (e.g. SORA~\cite{videoworldsimulators2024}, Pika~\cite{pika}, Gen-2~\cite{runaway}, and Dream Machine~\cite{luma}), and synthetic procedural modeling techniques (e.g., Kubric~\cite{greff2022kubric}, SceneX~\cite{zhou2024scenex}, SceneCraft~\cite{hu2024scenecraft}). The former generate sequences automatically but struggle with physical-correctness, shading and illumination, camera control, and temporal consistency. The latter can address the aforementioned issues but are usually for static 2D or 3D content, and do not address high-level storyboard control. 

Our key inspiration is to leverage LLM/VLM agents and 3D modeling engines to easily generate videos of realistic but synthetic content. While standard LLM/VLMs lack specific knowledge of the desired 2D/3D content and are missing  control of complex scene dynamics and camera motions, we use multiple collaborative LLM/VLM agents in an iterative process to overcome the shortcomings of prior methods. 

Our approach defines a pipeline consisting of three intertwined phases (Fig.~\ref{fig:frameworks}). The \textbf{Collaborative Generating Phase} feeds a user-provided natural-language-based video description to our \texttt{LLM-Director} Agent. This agent in turn produces instructions which are fed to our \texttt{LLM-Programmer} Agent, eventually yielding Python scripts for Blender resulting in a video containing the directed 2D/3D animation. The \textbf{Visual Reviewing Phase} uses a \texttt{VLM-Reviewer} Agent to provide feedback to the \texttt{LLM-Programmer} about the quality of the produced video sequence. This feedback causes the \texttt{LLM-Programmer} to iteratively alter its Python script output until successful generation. The \textbf{Library Updating Phase} automatically finds and/or updates the agent customized code libraries needed by Blender. All the agents use public tutorials and guidelines in a Retrieval Augmented Generation (RAG)~\cite{lewis2020retrieval} based framework to enhance specific knowledge. With our pipeline we have produced numerous videos, under simple natural-language based control, and produce video quality far superior to seven comparison baselines (e.g., Pika~\cite{pika}, Gen-2~\cite{runaway},  Dream Machine~\cite{luma}, and more -- see Sec.~\ref{sec:experiments}).

\noindent The contributions of our approach include:

\begin{enumerate}
\item {\textit{Collaborative Video Agents.}} The definition of multiple LLM/VLM agents that divide the phases of video generation into Director, Programmer, and Reviewer.

\item{\textit{Iterative Video Generation.}} The use of iterative reasoning and feedback loops to produce videos. This includes improving scene/video generation, the internal function calls, and the modeling/Blender scripts.

\item{\textit{3D Modeling Engine Use.}} Leveraging 3D modeling engine scripting for complex text-to-video generation with physical correctness, good shading and illumination, smooth camera motion, and temporal consistency as well as parameterized storyboard control.

\end{enumerate}
\section{Related Works}
\label{sec:related_works}


\subsection{Video Creation}


With booming of DM-based image synthesis~\cite{li2024training,chen2024parameter,hedocument2023,miao2025coeff,song2024refine,he2023globalmapper,he2023globalmapper,hecoho2024,hua2024finematch,patel2023deep}, video creation also attract dense efforts. Such end-to-end systems use text, image, and natural language descriptions to produce video clips of up to approximately a minute in length. The underlying engine can be diffusion based including, SORA~\cite{videoworldsimulators2024}, Stable Video Diffusion~\cite{blattmann2023stable}, VideoPoet~\cite{kondratyuk2023videopoet}, and PYoCo~\cite{ge2023preserve}. It can also use transformers like GenTron~\cite{chen2024gentron}, or use divide-and-conquer strategies for better control as Microcinema~\cite{wang2024microcinema}. Some may re-use existing videos~\cite{Peng_Chen_Wang_Lu_Qiao_2024}, or image-pose pairs for character animation~\cite{Ma_He_Cun_Wang_Chen_Li_Chen_2024, tang2025generative, gaussianstyle, song2024texttoon}. Others can be queued off audio~\cite{Yariv_Gat_Benaim_Wolf_Schwartz_Adi_2024} or can focus on particular video categories such as animating people~\cite{hu2024animate, 10446837, tang2025generative} or creating dances~\cite{zeng2024make}. However, these end-to-end generation systems struggle to maintain physically-correct motions and behaviors, proper shading and illumination, good camera control, and temporal consistency. Some systems address a few of these issues, such as ~\cite{bar2024lumiere} focus on temporal consistency and ~\cite{yang2024direct} provide separated object and camera controls.

\subsection{LLMs and Agents}
As inspiration for your work, we build upon the reasoning ability of LLMs and improved abilities provided by collaborating agents. As recently shown, LLMs are unexpectedly good at reasoning with chain of thoughts~\cite{wei2022chain, tang2024cardiff}. Moreover, they can be used to help with acting~\cite{yao2022react}, generating programming statements~\cite{chen2022program, wang2022self}, decision making~\cite{yao2024tree}, and performing iterative refinement~\cite{madaan2024self} without necessarily requiring fine-tuning or re-training~\cite{shinn2024reflexion}. In addition, the use of several collaborating agents has very recently found success in applications for various modalities~\cite{tang2024vidcomposition,liu2024empiricalanalysislargelanguage,kinmo}. For example, several agents, or components, collaborate for better planning~\cite{zhang2024combo}.

\subsection{Procedural and Synthetic Generation}
Our work also builds upon realistic procedural/synthetic scene generation.~\cite{raistrick2023infinite} produce static 3D scenes of the natural world. \cite{wang2024themestation} generate 3D assets based on exemplars.~\cite{deitke2022️} procedurally generate physically-enabled scenes. Kubric~\cite{greff2022kubric}, SceneX~\cite{zhou2024scenex}, and SceneCraft~\cite{hu2024scenecraft} use 3D modeling software (e.g., Blender) and LLMs to generate 3D static scenes. With these synthetic approaches, obtaining plausibly correct physical structures and good shading and illumination is much easier~\cite{bhatt2020design,zhang2022guided}.

However, these procedural outputs do not produce videos of the generated content and much less according to a user-guided storyboard. Enabling this additional step requires more complex high-level director control and feedback that are not provided previously.~\cite{yamada2024l3go} uses a trial-and-error approach to enable feedback but does not produce videos.~\cite{zhuang2024vlogger, liu2025contextual} use a diffusion-based approach to create video sequences with some notion of high-level director control, but are not synthetic content and thus suffers from aforementioned limitations of end-to-end video generation.

In our contrast, our approach is the first to support providing high-level storyboard using natural language, introduce multiple collaborating LLM agents to provide feedback and iterative improvement, and produce synthetic videos with good physical-correctness, shading and illumination, camera control, and temporal consistency.

\section{Method}
\label{sec:methods}

\begin{figure*}[!h]
  \centering
  \includegraphics[width=\linewidth]{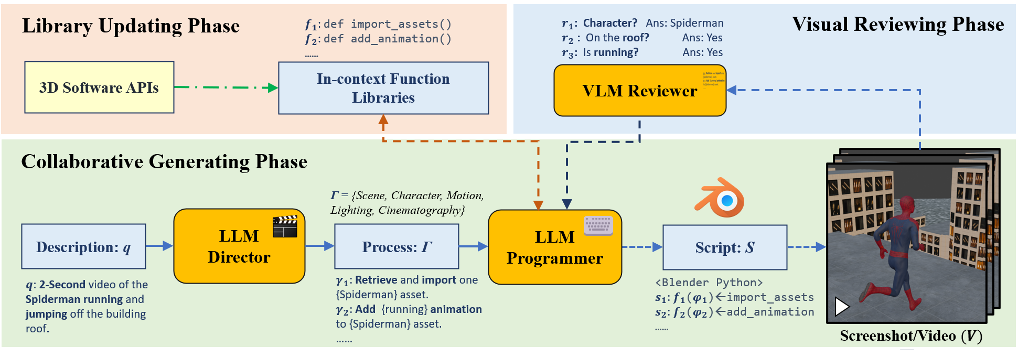}
  \caption{\textbf{Kubrick Framework.} Our video generation pipeline consists of multiple agent collaborations and iterative refinement of both generated visual results and code libraries. The \texttt{LLM-Director} decomposes the user-provided video description into functional sub-processes. Then, for each sub-process \texttt{LLM-Programmer} makes use of a set of function libraries to compose Python scripts for Blender. Intermediate screenshots and video outputs will be reviewed by \texttt{VLM-Reviewer} for visual quality and prompt-following evaluation. The review will be fed back to \texttt{LLM-Programmer} to iteratively improve the video. All three agents are improved by RAG based on public video tutorials and online documents.}
  \label{fig:frameworks}
\end{figure*}

Our multi-agent synthetic video generation framework (Fig.~\ref{fig:frameworks}) is divided into three main phases which we describe in turn. The system takes as input a text description of a desired video sequence and produces a set of scripts for 3D engine rendering, which ultimately results in a video/animation sequence. The pseudo-code is provided in Algorithm~\ref{alg:algorithm}.


\begin{algorithm}[tb]
\caption{Kubrick Video Generation}
\label{alg:algorithm}
\textbf{Input}: User-provided video description $q$.\\
\textbf{Parameter}: 3D asset and rigged armature collection $D$, domain knowledge of Blender video making $K$, list of agents $A$ = [\texttt{LLM-Director}, \texttt{LLM-Programmer}, \texttt{VLM-Reviewer}], RAG framework $RAG$, maximum review iteration $N_{max}$. \\
\textbf{Output}: Output video $V$
\begin{algorithmic}[1] 
\STATE Let $V=None$, $S=None$
    \STATE $A \leftarrow RAG(A, K)$ \hfill  \COMMENT{RAG initialization}
    \STATE $\Gamma \leftarrow  \texttt{LLM-Director} (q)$ \hfill  \COMMENT{Decomposition}

    \STATE $Lib \leftarrow  \texttt{LLM-Programmer} (\Gamma)$  
    
    \FOR{$\gamma_{i}$ in $\Gamma$} 
        \STATE $t = 0$, $r = None$

        \WHILE{$r$ and $t<N_{max}$}    
            \STATE $Lib \leftarrow \texttt{LLM-Programmer} (Lib, r)$ \COMMENT{Library Updating}
            \STATE  $s_{i} \leftarrow  \texttt{LLM-Programmer} (\gamma_{i}, Lib, r)$ \hfill  \COMMENT{Scripting}
            \STATE  $v_{i} \leftarrow  \texttt{Blender} (S + s_{i}, D)$
            \STATE $r \leftarrow \texttt{VLM-Reviewer}(\gamma_{i}, v_{i})$ \hfill \COMMENT{Visual Review}
            \STATE $t = t + 1$
        \ENDWHILE \hfill \COMMENT{Reviewing Loop}

        \STATE $S = S +s$
    \ENDFOR 
    \STATE $V \leftarrow \texttt{Blender} (S, D) $

\STATE \textbf{return} $V$
\end{algorithmic}
\end{algorithm}

\subsection{Collaborative Generating Phase}
\label{sec:generation}


In this first phase, a \texttt{LLM-Director} agent collaborates with a \texttt{LLM-Programmer} agent (as well as other phases). Designing and then creating a compelling video requires addressing many details. All-at-once solutions may easily lead to unreasonable results. Similar to the chain-of-thoughts~\cite{wei2022chain}, we observed that decomposing the task to sub-processes benefits video quality and text-to-video alignment. The decomposition may also contribute to effective visual reviewing and to reducing the number of generate-review-edit cycles. 

Taking inspiration from film making, the \textit{mise-en-sc\`{e}ne} (i.e., all aspects of the scene/video generation) is addressed by decomposing the generation process into the following five sub-processes~\cite{kawin1992movies}:

\begin{itemize}
    \item \textit{Scene}: The selection, texture, location, and scale of environmental 3D assets (e.g., trees, buildings).
    \item \textit{Character}: The selection, texture, location, and scale of the main actors (e.g., characters, people, animals).
    \item \textit{Motion}: The rigging, motion track, motion type, and moving speed of the 3D character assets (e.g. run, walk, jump off).
    \item \textit{Lighting}: The general lighting condition of the entire scene. (e.g., night, sunshine).
    \item \textit{Cinematography}: The camera parameters (e.g., FOV, position, resolution) and camera language (e.g., motion, vibration). 
\end{itemize}


\subsubsection{LLM Director.} 

We prompt the \texttt{LLM-Director} to sequentially focus on the aforementioned five sub-processes. Specifically, given an initial user provided description $q$, the \texttt{LLM-Director} decomposes the description into text guidance for each of the sub-processes $\Gamma$ as in the following equation: 

\begin{equation}
\label{eq:director}
    \Gamma = \{\gamma_{i} | i=1, 2, ..., n\} \leftarrow  \texttt{LLM-Director} (q)
\end{equation}

\begin{figure}[tb]
  \centering
  \includegraphics[width=\linewidth]{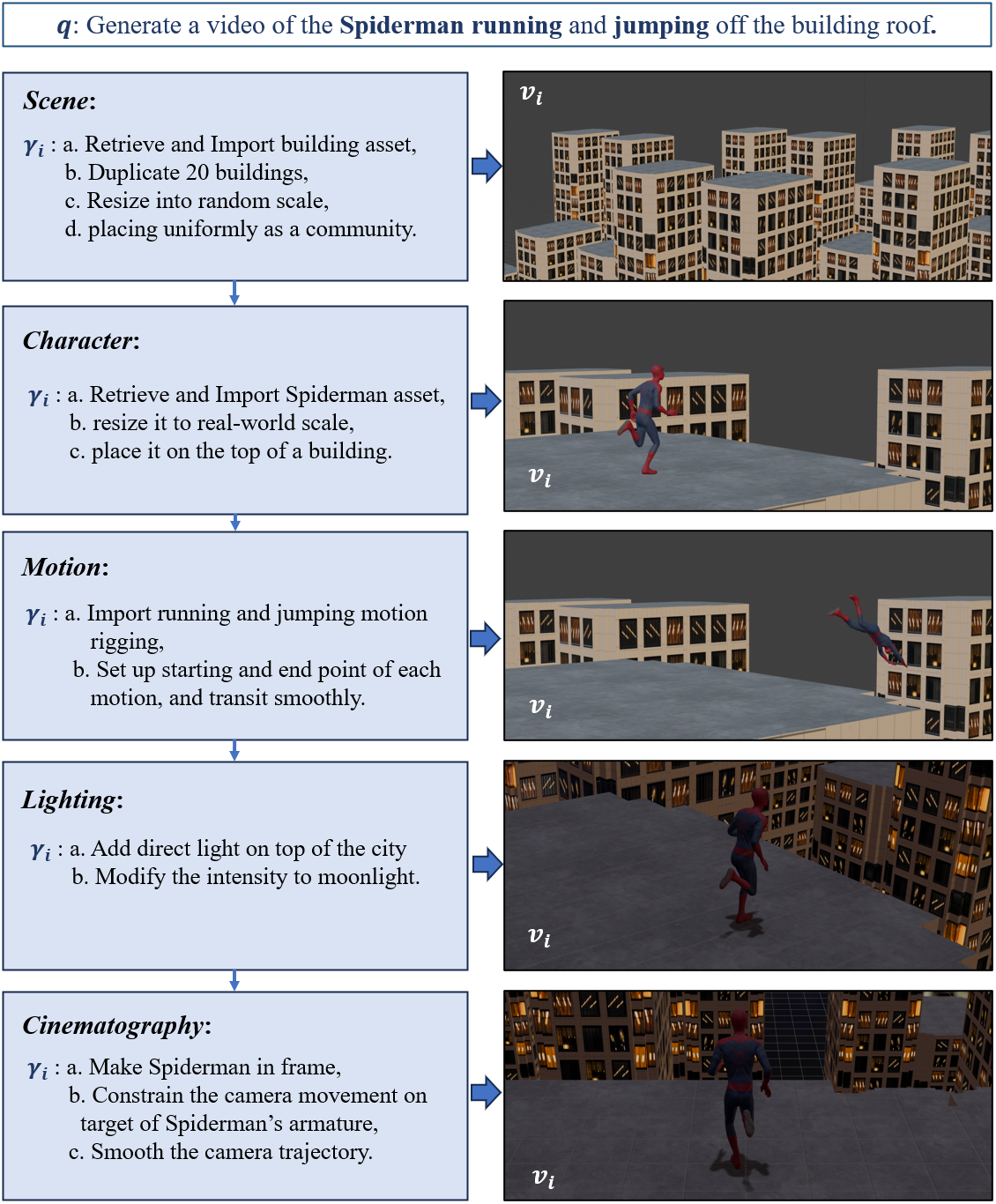}
  \caption{\textbf{Collaborative Generation.} The video description $q$ is decomposed by \texttt{LLM-Director} to five sub-processes of \textit{Mise-en-scène}. Each sub-process handles specific tasks during synthetic video generation. Each sub-process description $\gamma_{i}$ guides scripting by \texttt{LLM-Programmer} to render corresponding intermediate visual output $v_{i}$.} 
  \label{fig:decomposition}
\end{figure}

As Fig.~\ref{fig:decomposition} shows, each sub-process addresses clearly distinct aspects of the video. Since user descriptions often lack details, the decomposition facilitates the \texttt{LLM-Director} to complete the complex details of 3D scene synthesis and video dispatching descriptions for each sub-process. This enriching of each sub-process is crucial for completeness as well as diversity. Moreover, this decomposition benefits the visual and reasoning ability of the \texttt{VLM-Reviewer}.

\subsubsection{LLM Programmer} 

In sequence, each sub-process $\gamma_{i}$ is converted by \texttt{LLM-Programmer} to an intermediate code snippet $s_{i}$. Once all code snippets have been completed, they are combined to form the final script $S$:

\begin{equation}
\label{eq:programmer}
    S = \{s_{i} | i=1, 2, ..., n\} \leftarrow  \texttt{LLM-Programmer} (\Gamma)
\end{equation}
 
Each code snippet $s_{i}$ will be rendered by a 3D modeling engine (i.e., in our case we use Blender) to intermediate visual output $v_{i}$. Each code snippet $s_{i}$ consists of selected function calls $f_{i}$ given arguments $\varphi_{i}$. It can be represented as $s_{i}=\{f_{j}(\varphi_{j}) | f \in Lib \}$, where $f$ is selected from possible functional call library database $Lib$. Our goal is to search for a good selection of both $f_{j}$ and $\varphi_{j}$ given sub-process $\gamma_{i}$ with the iterative refinement help of visual reviewing and library updating phases, in order to ensure good visual quality and text-to-video alignment. The accumulated final script $S$ is utilized for creating the final video $V$. 

\subsection{Visual Reviewing Phase}
\label{sec:visual reflection}
The \texttt{VLM-Reviewer} coordinates visually-based reviewing. This phase scores each intermediate visual output $v_{i}$ and thus (indirectly) the accumulated final output $V$. The system requests the \texttt{VLM-Reviewer} to evaluate both prompt-following and visual quality of $v_{i}$ of each sub-process $\gamma_{i}$. The ultimate goal is to improve code snippet $s_{i}$'s score. The optimization can be defined as: 
 
\begin{equation}
    S \leftarrow \underset{s_{i}}{\argmax} \sum_{i=1}^{n} \texttt{VLM-Reviewer} (\gamma_{i}, v_{i})
\end{equation}

In particular, the score from \texttt{VLM-Reviewer} is encoded in text format as bulletin reviews $r$ to feed back to the \texttt{LLM-Programmer} (e.g., "The object is flying in the sky instead of standing on the ground"). In addition to the reviews, \texttt{VLM-Reviewer} also suggests several improvements to \texttt{LLM-Programmer} (e.g., "change color", "change motion"). Then, the \texttt{LLM-Programmer} will refine current code snippet from $s_{i}$ to $s_{i}'$, given the available function set $Lib$, the review $r$, and sub-task $\gamma_{i}$:

\begin{equation}
    s_{i}' \leftarrow \texttt{LLM-Programmer} (\gamma_{i}, Lib, r)
\end{equation}

\begin{figure}[tb]
  \centering
  \includegraphics[width=\linewidth]{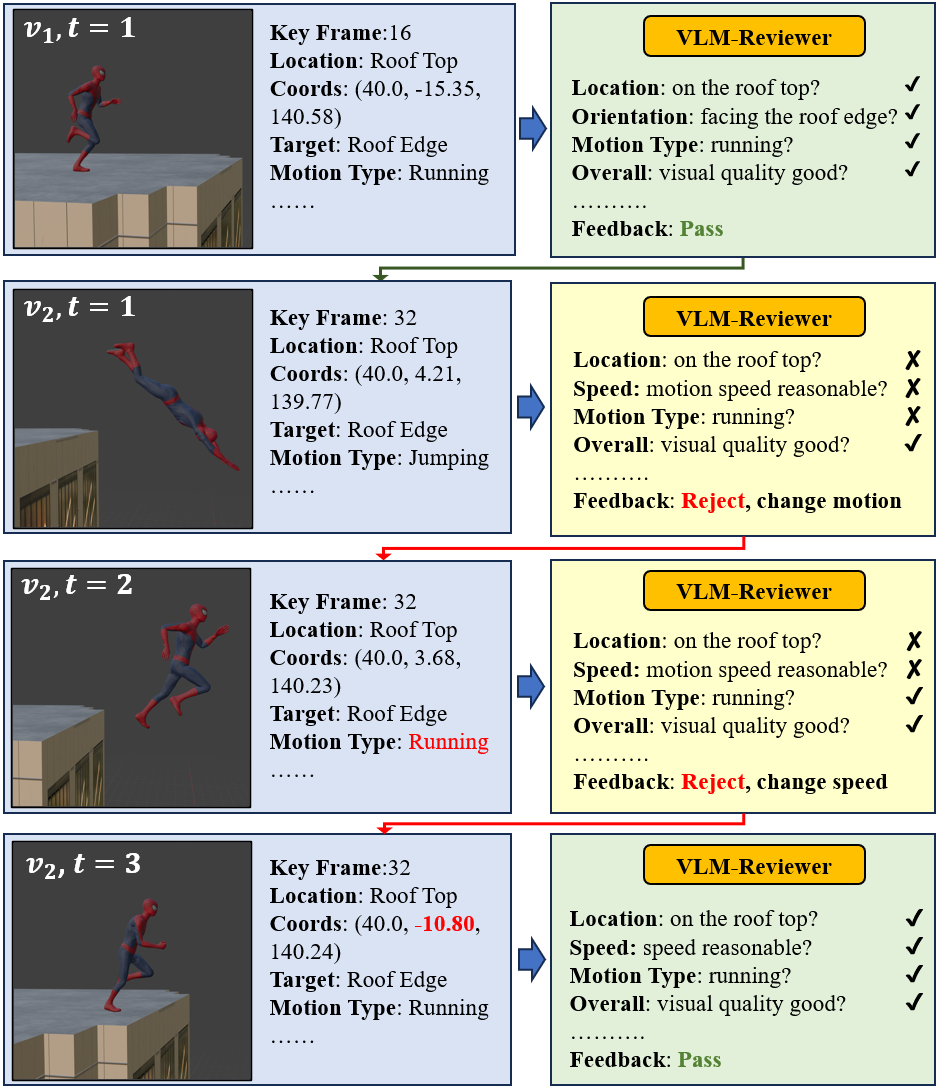}
  \caption{\textbf{Dynamic Content Evaluation.} The \texttt{VLM-Reviewer} evaluates a set of comprehensive metrics (e.g. character location coordinates, motion type, overall quality) to assess the current intermediate visual output. The metric set is decided by the \texttt{VLM-Reviewer} for each sub-process. The feedback for iteration $t$ will be used to improve the scripts for iteration $t+1$. A similar evaluation is conducted for camera motion, and the correctness of the 3D scene in the background.}
  \label{fig:evaluation}
  \vspace{-1em}
\end{figure}

\subsubsection{Dynamic Content Evaluation}
Unlike former works which evaluate static 3D scene generation~\cite{blendergpt,hu2024scenecraft}, we evaluate character motions and camera motions. As Fig.~\ref{fig:evaluation} shows, for the character motion the \texttt{LLM-Programmer} fixes the camera in a location so as to keep the main character 3D assets in frame and in focus. Since the character may have multiple motions, the trajectory of each motion is evaluated separately. Specifically, the \texttt{VLM-Reviewer} samples the starting and ending coordinates of the character's armature for each motion, text-based location coordinates, motion type, motion speed, and potentially other metrics. In addition, the \texttt{VLM-Reviewer} is provided with screenshots of key frames. Based on the answers of all the evaluation questions, the \texttt{VLM-Reviewer} will give a overall evaluation of "Pass" or "Reject", together with improvement suggestions for the rejected questions. If $v_{i}$ is rejected at iteration $t$, corresponding feedback will be given to \texttt{LLM-Programmer} for the scripting improvement in iteration $t+1$. For example, if the ending coordinates of a motion are not reasonable, \texttt{LLM-Programmer} will take the feedback to modify the ending point arguments ($\varphi_{i}$) when calling corresponding motion function $f_{i}$. Finally, all scripts will be accumulated for each sub-process and used to render a full-length video. 

The evaluation of the camera motion is with a similar logic but with denser key-frame sampling to ensure the smoothness of the camera movement. In practice, video descriptions that indicate significant movement of either character or camera (but not both) will have better success rate.

\subsection{Library Updating Phase}
In addition to function argument $\varphi_{i}$ refinement, addressing more complex feedback may require fundamental improvement of the function itself; e.g., if the current library $Lib$ has the motion assign function of a single motion, but the user description indicates multiple sequential motions. The visual reviewer feedback may ask \texttt{LLM-Programmer} to compose a necessary new function to temporally align multiple motions (Eq.~\ref{eq:library update}). Typically, the modification of $Lib$ includes replacing, adding, or removing the current function libraries. Modifications may not be necessary for minor refinement cases, and \texttt{LLM-Programmer} handles the implementations. For complex $q$, the updated $Lib$ further improves the performance than refinement $\varphi_{i}$ alone.

\begin{equation}
\label{eq:library update}
    Lib \leftarrow \texttt{LLM-Programmer} (Lib, r)
\end{equation}

\subsection{Retrieval-based Instructive Learning}

The latest LLMs are well-known for their coding capability in popular programming languages. However, we make use of Retrieval Augmented Generation (RAG)~\cite{lewis2020retrieval} to improve performance in synthetic video generation and Blender-oriented scripting. To this end, we use Blender API documents to improve the API calling correctness of the \texttt{LLM-Programmer} and reduce the number of feedback loops. In addition, to improve the \texttt{LLM-Director} and \texttt{VLM-Reviewer} we use online video tutorials (typically made by Blender experts) related to the user-supplied video description. For each video, we use the provided video subtitles (e.g., English text). We setup the retrieval database after receiving the user-provided video description. Then, RAG is used during the execution of each agent.

\begin{figure*}
  \centering
  \includegraphics[width=0.95\linewidth]{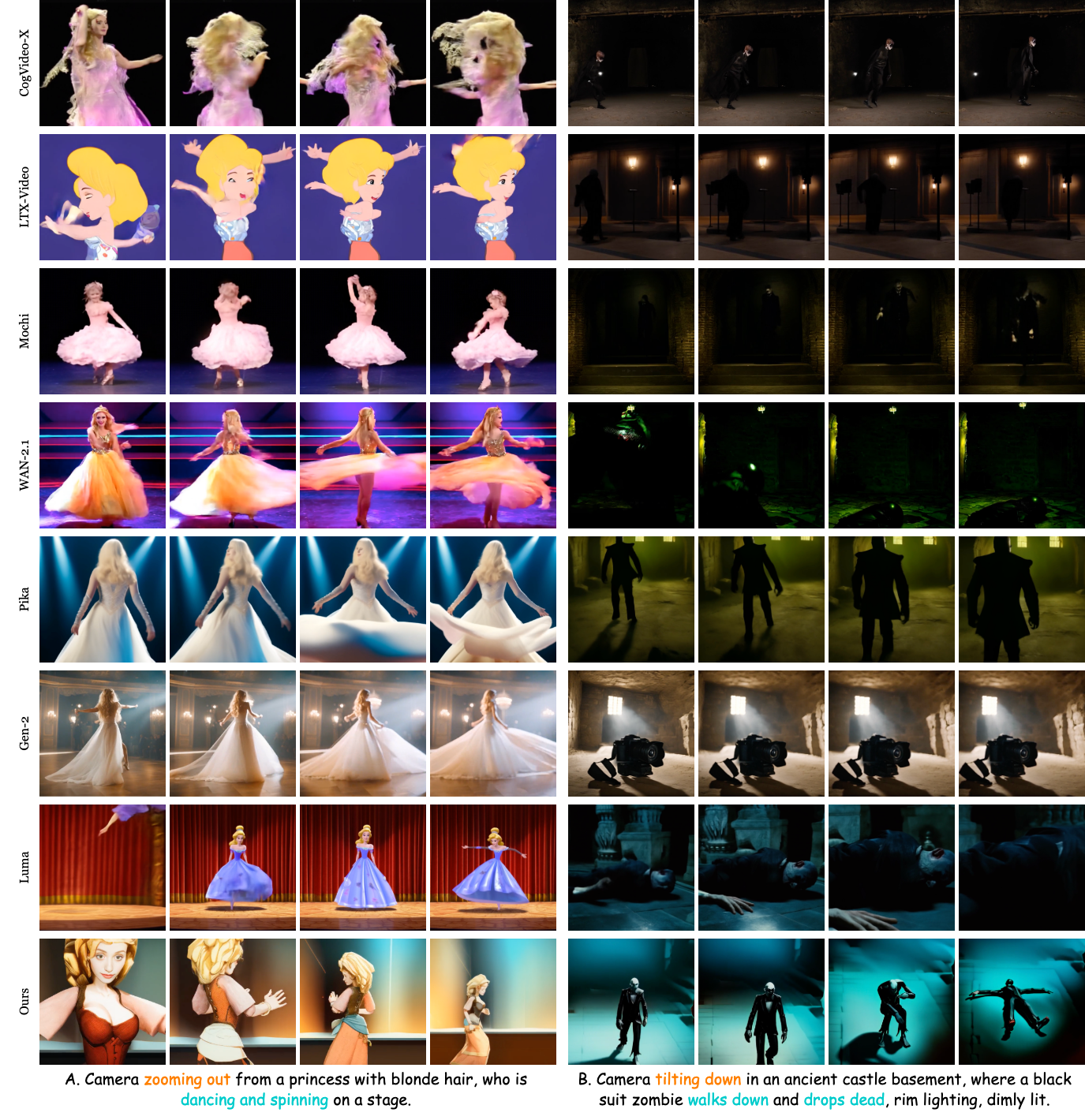}
  \centering
  \caption{\textbf{Qualitative Comparisons.} Our method excels in handling character/object motions (highlighted in \textcolor{cyan}{blue}) and complex camera moves (highlighted in \textcolor{orange}{orange}). Pika, Gen-2, and LTX-Video struggle to generate reasonable character motions. While WAN-2.1, Mochi, and CogVideo-X can synthesize character movement, their outputs are heavily distorted and contain significant artifacts. Additionally, all baseline models exhibit limited camera motion. Gen-2 misinterprets camera motion prompts as the main character (right), while Luma fails to accurately follow character motion. See the appendix for more results.
  }
  \label{fig:qual1}
\end{figure*}


\section{Experiments}
\label{sec:experiments}

\newcommand{\datasetname}{21-video-dataset }




\subsection{Dataset and Implementation Details}

We create 21 example text-based video descriptions to use as input to our method and to the baselines. For those descriptions, we create a 3D resource dataset, \textit{Odyssey}. The dataset consists of character and scene object 3D assets, and rigged armatures which can be implemented on all characters. We collect 3D assets (with textures) from Sketchfab~\cite{sketchfab} and rigged armatures from Mixamo~\cite{mixamo}. For RAG database, we collect the subtitles of 261 Blender videos from public video sharing platform, together with official Blender API documentations~\cite{Blenderapi}. Then, using the \textit{Odyssey} dataset and our method, we generate 21 example videos. 

Our pipeline needs online access to GPT-4V~\cite{OPENAI} with RAG knowledge~\cite{ragOPENAI} and a working version of Blender. As a side note, our pipeline does not prescribe the length of generated videos. Rather, our storyboarding process enables variable length videos as well as assembling longer videos. However, as videos get longer more review iterations are likely needed. We found a good compromise to be $N_{max}=15$ in our tested videos.


\subsection{Evaluation Metrics}

We evaluate videos using five metrics covering visual quality, fidelity, and description-following ability. For overall visual quality and fidelity, we implement frame-wise Frechet Inception Distance (FID)~\cite{heusel2017gans}, Frechet Video Distance (FVD)~\cite{unterthiner2018towards}, and video quality assessment (VQA) score~\cite{wu2023exploring}. Since FVD and FID evaluate the similarity between our generated videos to high-quality video references, we use the synthetic movie Sintel~\cite{Sintel} as our reference. Sintel is a manually crafted Blender-generated movie and considered to be of professional quality. We randomly sample 21 video clips from the Sintel movie (4 to 5 seconds each). Then we sample 21 same-length video clips from our generated videos for comparisons. Moreover, we use CLIP Similarity (CLIP-SIM)~\cite{wu2021godiva} as the description-following metric. It evaluates the semantic similarity between the generated videos and text descriptions.

\begin{table}[t]
    \small
    \centering
    \caption{\textbf{Quantitative results.} We report visual metrics of generated videos based on same set of input text prompt across 7 baselines. Our framework shows the best overall performance of visual quality (FVD, FID) and prompt following (CLIP-SIM). Our VQA score is also competitive to the close-sourced baseline Luma. Best results are in \textbf{bold}, second-best results are \underline{underlined}.}

    \begin{tabular}{c|c|c|c|c}
        \toprule
        Method & FVD $\downarrow$ & FID $\downarrow$ & VQA $\uparrow$ & CLIP-SIM $\uparrow$ \\
        \midrule
        CogV-X  & 2783.48 & 302.71 & \textbf{51.10} & 0.1994 \\
        Wan-2.1 & 3088.93 & 307.68 & \underline{48.14} & 0.2021 \\
        LTX & 3174.03 & 306.70 & 17.79 & 0.1986 \\
        Mochi & 2999.18 & 310.23 & 33.33 & 0.1948 \\
        
        Gen-2  & 3463.46 & 298.82 & 30.59 & 0.2042 \\
        Pika  & 2885.68 & 298.21 & 14.04 & 0.2044 \\
        Luma & \underline{2728.90} & \underline{261.24} & 40.97 & \underline{0.2068} \\

        Ours & \textbf{2691.44} & \textbf{245.43} & 40.31 & \textbf{0.2155} \\
        \bottomrule
    \end{tabular}
    \label{tab:quan}
\end{table}


        


        


\subsection{Quantitative Comparisons}
\label{sec:quan}


 We compare our model against seven baseline models: Gen-2 \cite{runaway}, Pika \cite{pika}, Luma Dream Machine \cite{luma}, CogVideo-X~\cite{yang2024cogvideox}, Wan-2.1~\cite{wan2024video}, Mochi~\cite{mochi2024}, and LTX-Video~\cite{hacohen2024ltxvideo}. 21 videos are generated by each model using the same set of text descriptions. For fairness, each video is only generated once without cherry-picking.

Quantitative comparisons are shown in Tab.~\ref{tab:quan}. Our model outperforms all baseline models in three metrics and shows competitive performance on VQA score compared to the second best overall close-sourced model Luma. The latest open-sourced model CogVideo-X only outstands on VQA score but regresses on all other metrics. It demonstrates that our generated videos keep the balance of visual quality $\&$ fidelity (FVD and FID), and better prompt following indicated by higher CLIP-SIM score (see Fig.~\ref{fig:qual1}).


\begin{table}[t]
\centering
\caption{\textbf{User Study}. We compare different text-to-video generation methods in four criteria. Each user will watch 2 random videos generated by each model, and rate the score of each criteria in the scale from 1 to 5. Below shows the average score of each criteria. Our framework shows the best overall performance on all criteria by a significant margin. Best results are in \textbf{bold}, second-best results are \underline{underlined}.}
\label{tab:user_study_no_std}
\setlength{\tabcolsep}{4pt}
\small
\begin{tabular}{lcccc}
\toprule
\multirow{2}{*}{\textbf{Method}} & \multicolumn{4}{c}{\textbf{Evaluation Criteria}} \\
\cmidrule(lr){2-5}
 & Prompt $\uparrow$ & Motion $\uparrow$ & Camera $\uparrow$ & Overall $\uparrow$ \\
\midrule
CogV-X & 2.95 & 2.43 & 2.47 & 2.52 \\
Wan-2.1 & \underline{3.48} & \underline{3.28} & 3.30 & \underline{3.35} \\
LTX & 2.27 & 1.85 & 2.22 & 1.87 \\
Mochi & 3.47 & 2.88 & 3.07 & 3.13 \\
Pika & 2.97 & 2.65 & 2.85 & 2.87 \\
Gen-2 & 2.67 & 2.38 & 2.60 & 2.52 \\
Luma & 3.47 & 3.28 & \underline{3.35} & 3.32 \\
\textbf{Ours} & \textbf{4.18} & \textbf{3.83} & \textbf{4.02} & \textbf{3.93} \\
\bottomrule
\end{tabular}
\label{tab:user_study}
\end{table}

\subsection{Qualitative Comparisons}
\label{sec:qual}



\paragraph{Evaluation Visualization} Fig.~\ref{fig:qual1} shows visual comparisons on text descriptions. We provide additional visual results in the appendix Fig.~\ref{fig:supp_qual2} as C and D. Although Pika and Gen-2 \cite{runaway} show natural and realistic lighting, they fail to synthesize highly dynamic motions and only show minor motion flows between key frames. Given complex descriptions, Gen-2 misunderstands camera control prompts as the main character (in C), and the results generated by Luma fail to follow the character motion prompts (in A and B). Although CogVideo-X and WAN-2.1 produce high-resolution and more realistic videos, they still struggle with large motion patterns, leading to significant blurriness. All baseline models are prone to visual artifacts, particularly in arms and garments, and suffer from motion inconsistency. In contrast, our approach achieves greater motion magnitude, precise camera control, and superior temporal consistency.


\paragraph{User Study}
In addition, we conducted a user study to compare the seven baseline video generation methods with ours. 30 users
participated in our study. For each video description, we randomly sample 2 videos from our generated videos and from each of the baseline methods (a total of 16 videos). The user is first presented with a video description and then watches the corresponding video. The order of the videos is random and the method used is hidden from the user. The user is asked to rate the video on a scale of 1 to 5 for each of four criteria: (1) prompt following, (2) character motion stability, (3) camera motion quality, and (4) overall quality. We report the average score for each model and for each criteria in Tab.~\ref{tab:user_study}. The table shows that our method significantly outperforms all baseline models. It also highlights the superiority of our model in following the character motion and camera control instructions embedded in the video text description. More details are provided in the Supplementary Fig.~\ref{fig:supp_user_study}.

\begin{figure}
\centering
\includegraphics[width=0.95\linewidth]{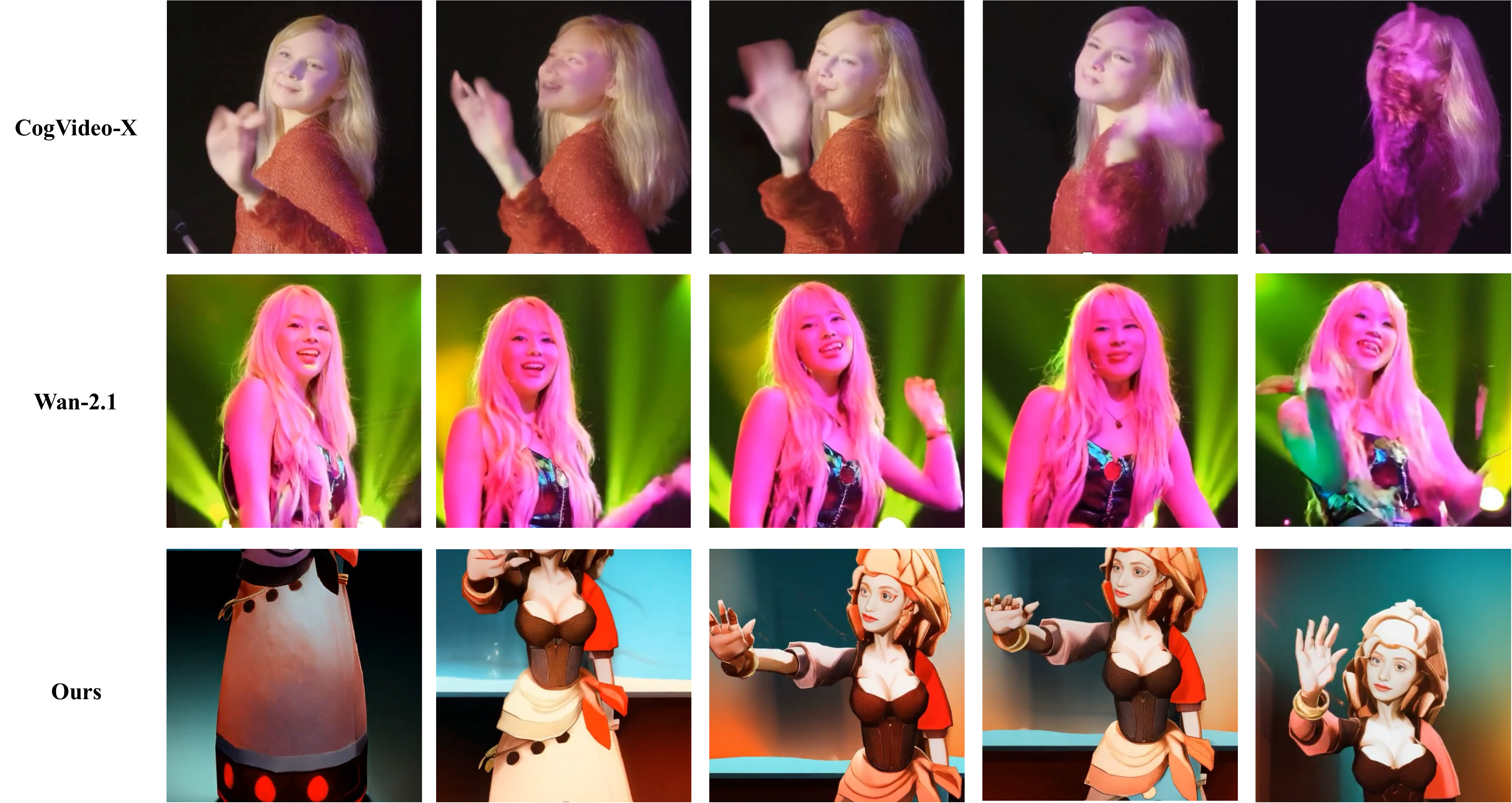}
\caption{\textbf{Motion and Camera Analysis.} We prompt the system with ``A girl waving her hands with camera tilt up." Compared to existing generation frameworks, our approach produces more diverse character motions and dynamic camera movements. Additionally, current methods often struggle with large and rapid motion patterns, leading to noticeable blurring artifacts.}
\label{fig:motion}
\end{figure}

\paragraph{Analysis}
Compared to existing end-to-end text-to-video frameworks, our LLM-based approach demonstrates significantly improved character motion quality and more dynamic camera control. As shown in Fig.~\ref{fig:motion}, our method does present motion blur, which is a common issue in traditional models when handling large or fast movements. By leveraging the reasoning capabilities of LLMs and agent-based framwork, our system generates a wider variety of motion patterns, leading to more natural and expressive character animation. Additionally, our approach enhances camera motion control, enabling smoother and more intentional movements. These improvements collectively contribute to a more visually coherent and aesthetically pleasing motion synthesis, making our method a strong alternative to conventional text-to-video generation techniques.

\subsection{Ablation Studies}
\paragraph{RAG Contributions} The Blender scripting ability of our agents is limited by the cutoff date of GPT-4V. We observed that RAG based on the latest Blender documents notably improves the API calling correctness. To further improve the ability to process complex video descriptions, we make use of text (e.g., subtitles) extracted from video tutorials. These tutorials provide short-cut tips, knowledge of Blender geometric nodes and plugins to help agents quickly connect complicated prompts to API calling combinations. Altogether, RAG reduces the average iterations of visual reviewing and library updating from 20.25 to 11.90.

\paragraph{Agent Assignment} The selection and number of agents affect the overall performance. We tried several agent configurations but found the proposed setup to be the simplest and most effective. For the \texttt{LLM-Programmer}, we experimented with further decomposition of this agent into specific sub-agents (e.g., Gaffer, Camera man, Scene designer). However, no explicit performance improvement is observed. We also experimented with expanding the visual review loop to involve \texttt{LLM-Director}, but its sub-process outputs did not change significantly. Thus, during reviewing we only iteratively refine library and scripts.

\paragraph{LLM/VLM Selection} We tested Claude-3-Sonnet~\cite{Claudev3} as the alternative backbone for all agents and also used LLaMa-3-70B-Instruct~\cite{dubey2024llama} as the alternative backbone of \texttt{LLM-Director/Programmer}. However, in all cases, using GPT-4V excels in coding and visual understanding. For example, using GPT-4V reduces in average 34.79\% reviewing iterations compared to using other LLM/VLMs.


\subsection{Limitations}
\texttt{VLM-Reviewer} may generate incorrect understanding of subtle 3D visual information, like character orientation, and action type. Visual review quality limits efficiency and overall quality. In practice, minor human-in-the-loop involvement will increase the successful rate of video generation. 

\section{Conclusion}
\label{sec:conclusion}
End-to-end video generation has been struggling with physical correctness, camera control, and temporal consistency. While procedural approaches require tedious manual crafting, our method successfully leverages LLM/VLM agents and 3D engines (e.g. Blender) for automatic synthetic video generation. We design collaborative generation, visual reviewing, and library updating phases as part of a robust and efficient pipeline, and use RAG with video making and 3D engine knowledge for further improvement. Our approach outperforms seven commercial or open-sourced video generation models in multiple metrics, and for the ability of prompt following, character motion, camera motion, and visual fidelity as per our user study.



{
    \small
    \bibliographystyle{ieeenat_fullname}
    \bibliography{main}
}

\newpage
\setcounter{figure}{0}
\setcounter{table}{0}
\setcounter{section}{0}
\clearpage
\setcounter{page}{1}
\maketitlesupplementary

\section{Generated Videos}
In Supplemental Fig.~\ref{fig:supp_qual}, we provide seven additional video clips as in Fig.1 (main paper). The generated videos show a good ability of instruction-following, consistent character motion, and smooth camera controlling.

\begin{figure*}
  \centering
  \includegraphics[width=0.91\linewidth]{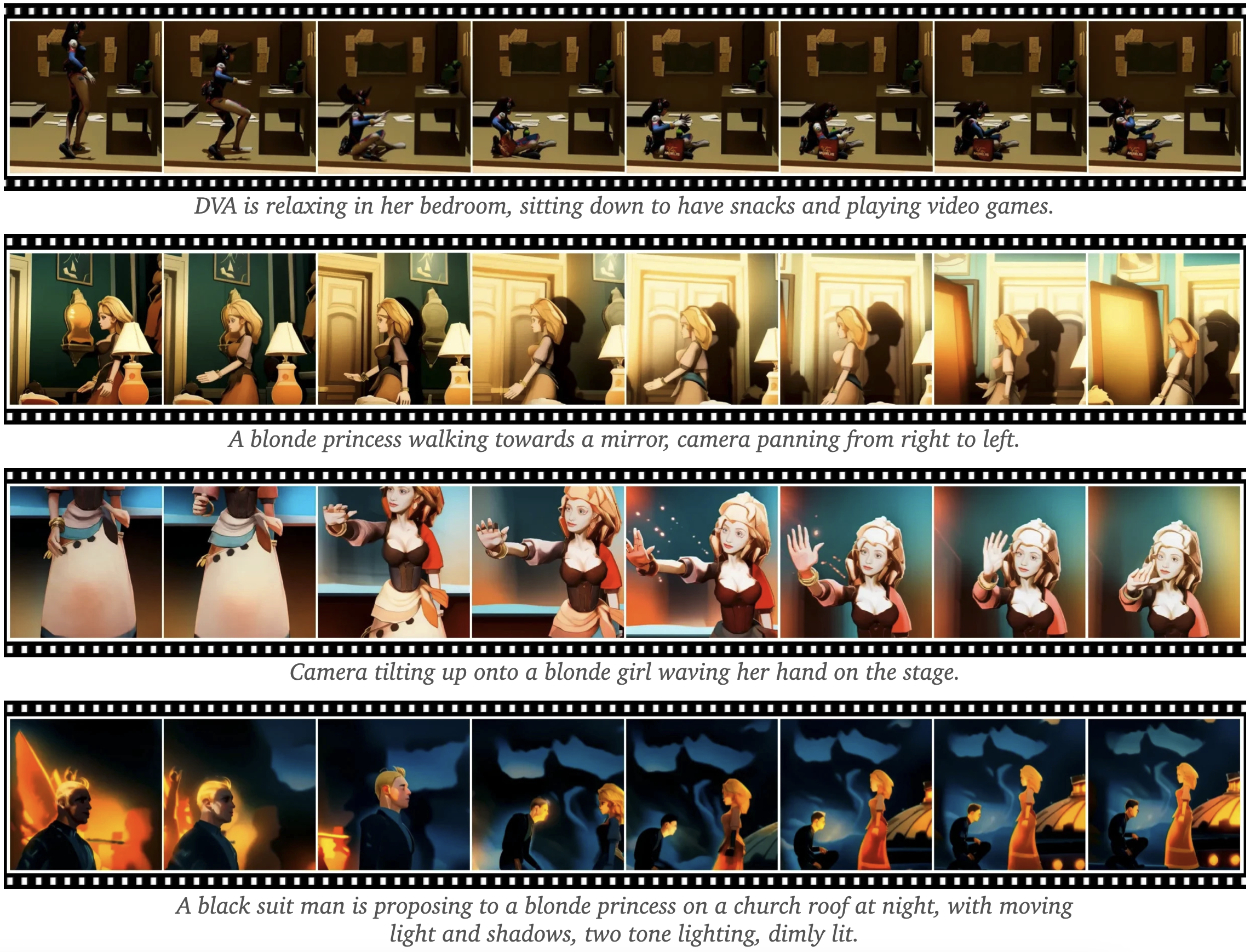}
  \includegraphics[width=0.91\linewidth]{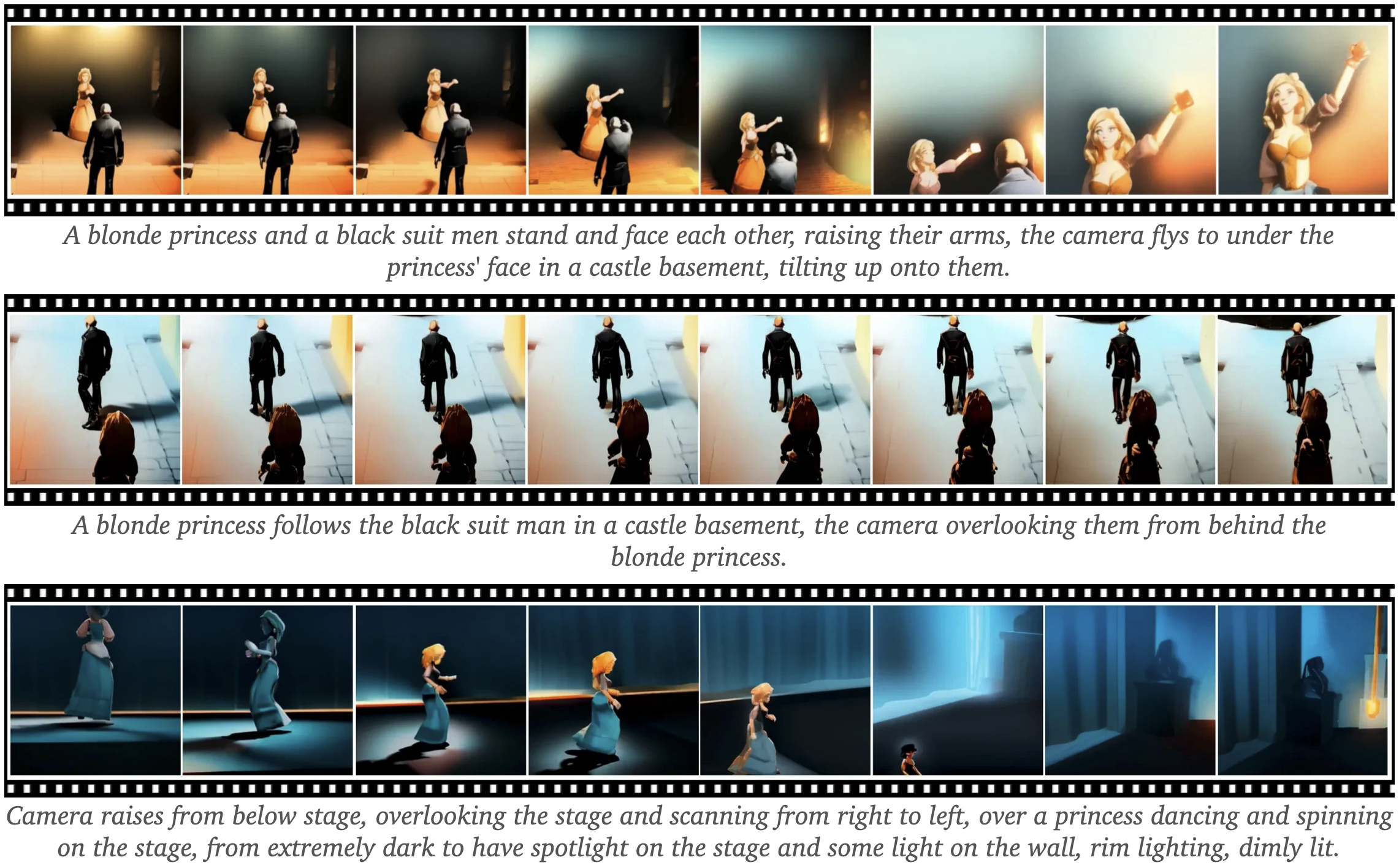}
  \centering
  \caption{\textbf{More Generated Video Clips.} The corresponding text prompts are provided below each video clip.}
  \label{fig:supp_qual}
\end{figure*}

\section{Additional Qualitative Comparisons}

\begin{figure*}
  \centering
  \includegraphics[width=0.95\linewidth]{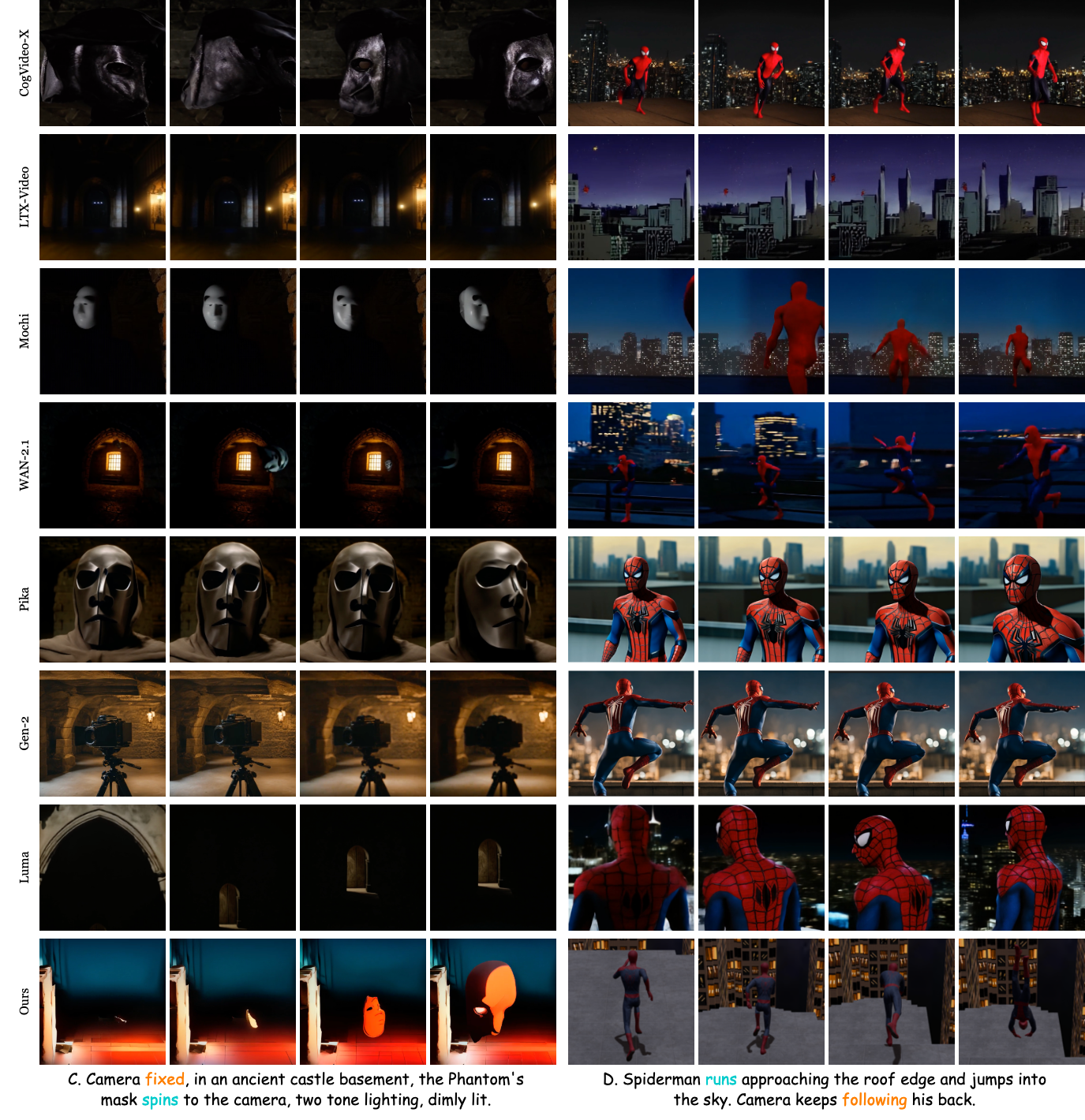}
  \centering
  \caption{\textbf{More Qualitative Comparisons.} Our method excels in handling character/object motions (highlighted in \textcolor{cyan}{blue}) and complex camera moves (highlighted in \textcolor{orange}{orange}).}
  \label{fig:supp_qual2}
\end{figure*}

This part is a further extension of the qualitative comparison in the main paper. The visual comparisons are shown in Fig.~\ref{fig:supp_qual2}. The results of the baseline models in C suggest that prompts of a fixed camera always lead to more static or less dynamic motions of the objects. Our agent-based framework does not present this issue. For the example in D, our method presents higher motion dynamics over existing frameworks.

\section{Video-to-Video Style Transfer}
\begin{figure*}
  \centering
  \includegraphics[width=\linewidth]{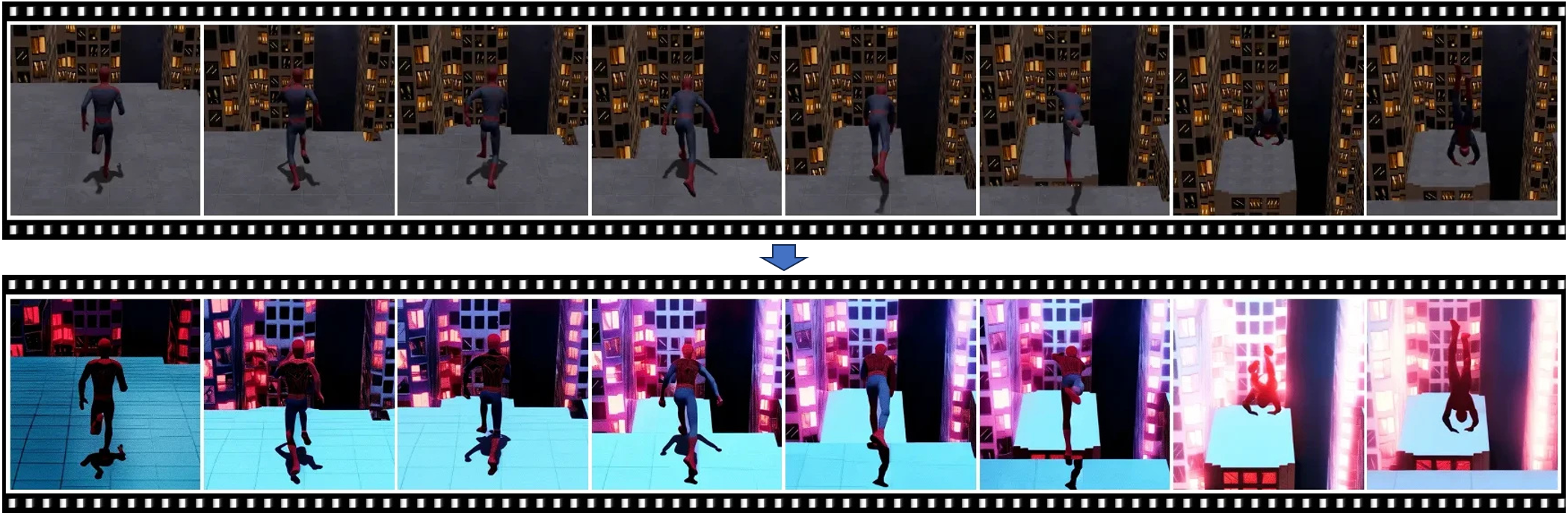}
  \centering
  \caption{\textbf{Video-to-Video Style Transfer.} We use AnimateDiff~\cite{guo2023animatediff} and ControlNet~\cite{zhang2023adding} (given both canny edge and depth) for each key frame output by our method. The converted video (lower row) inherits consistent character motion and smooth camera motion from the video generated by our method (upper row). It shows the potential of using our generated videos as prior for diversely-styled video generation. Note that the color inconsistency across key frames results from the randomness in the diffusion model.}
  \label{fig:supp_style_transfer}
\end{figure*}

In Fig.~\ref{fig:supp_style_transfer}, we show video-to-video style transfer by applying the diffusion-based model AnimateDiff~\cite{guo2023animatediff} and ControlNet~\cite{zhang2023adding} (given both canny edge and depth) to each key frame output of our method. It transfers our input images into the altered animation style. The converted video inherits consistent character motion and smooth camera motion from the video generated by our method. It shows the potential of using our generated videos as prior for style-varying video generation.

\section{Prompts for Agents}
\begin{figure*}
  \centering
  \includegraphics[width=0.9\linewidth]{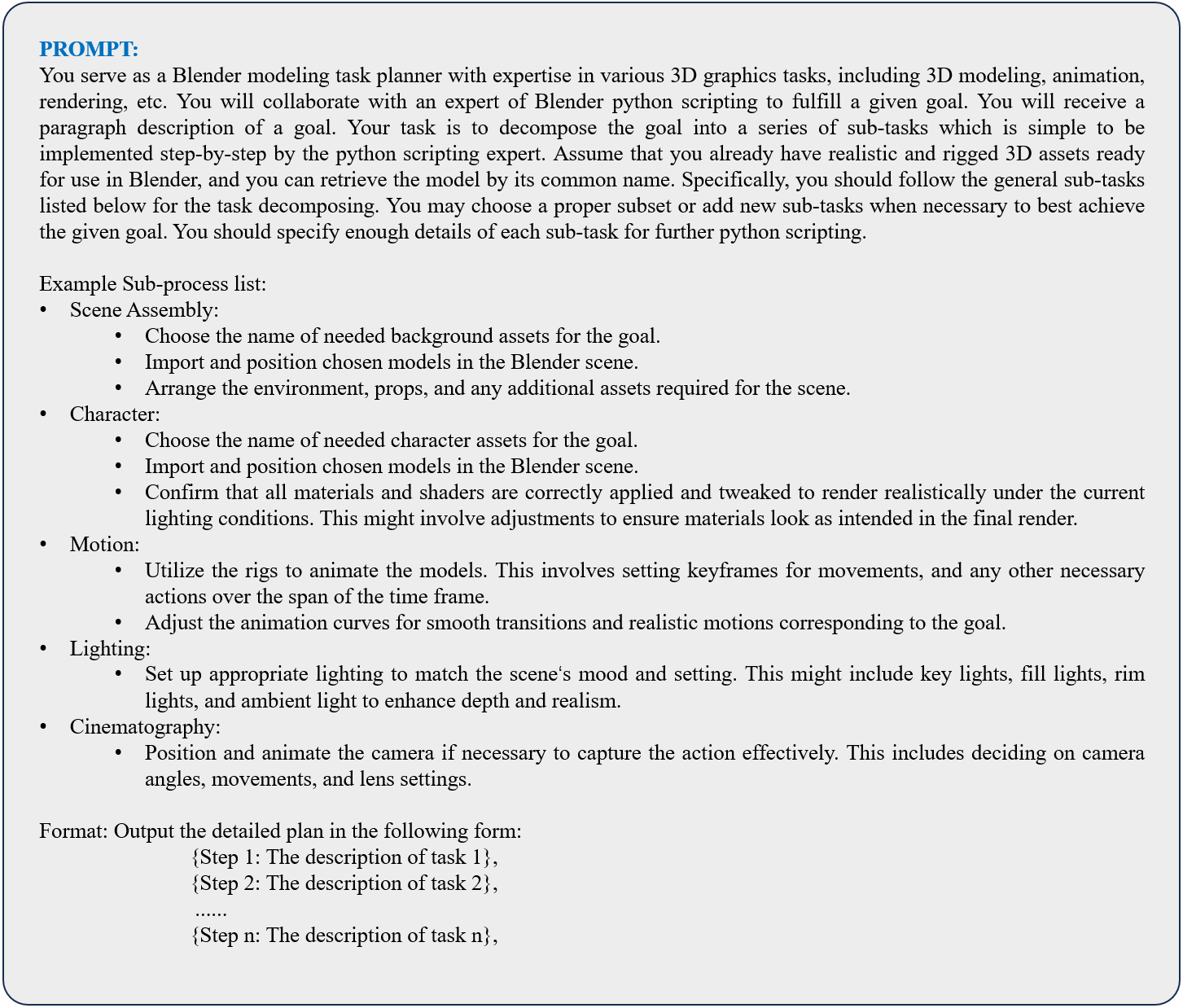}
  \centering
  \caption{\textbf{Prompt for Sub-process Decomposition.} An example text prompt for \texttt{LLM-Director} sub-process decomposition.}
  \label{fig:supp_prompt_director}
\end{figure*}

\begin{figure*}
  \centering
  \includegraphics[width=0.9\linewidth]{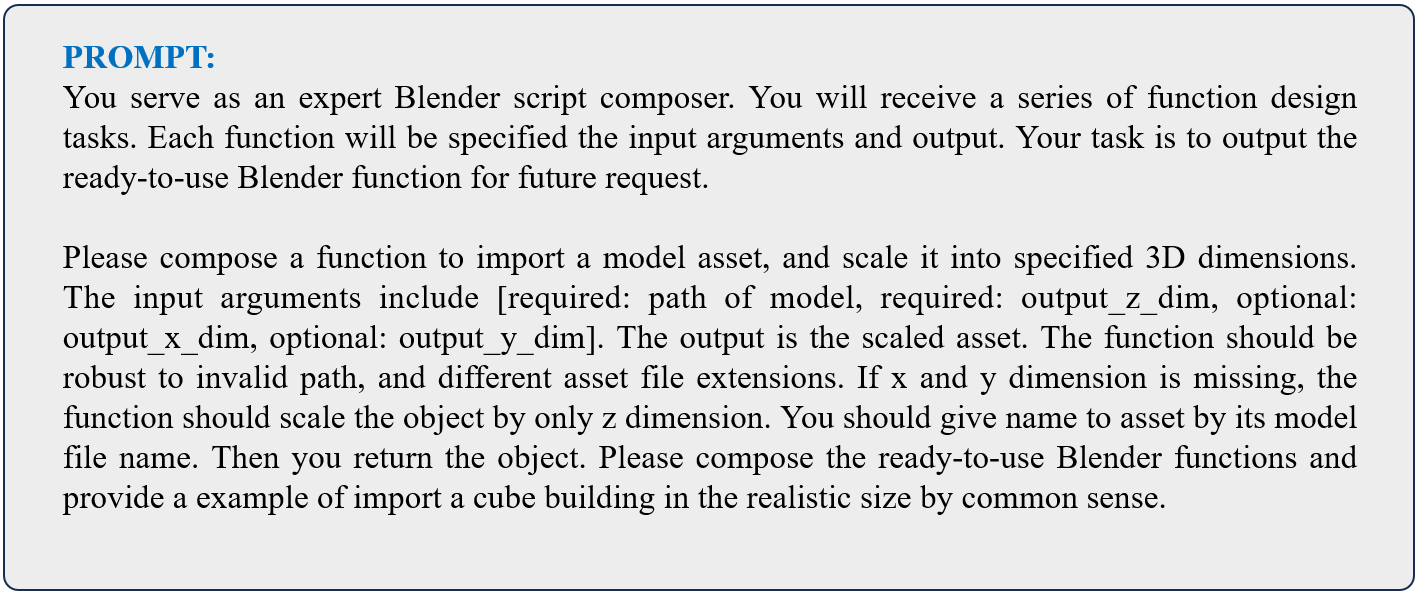}
  \centering
  \caption{\textbf{Prompt for Library Function Generation.}  An example text prompt for \texttt{LLM-Programmer} to compose a function for importing and scaling of a 3D asset.}
  \label{fig:supp_prompt_programmer}
\end{figure*}

\begin{figure*}
  \centering
  \includegraphics[width=0.9\linewidth]{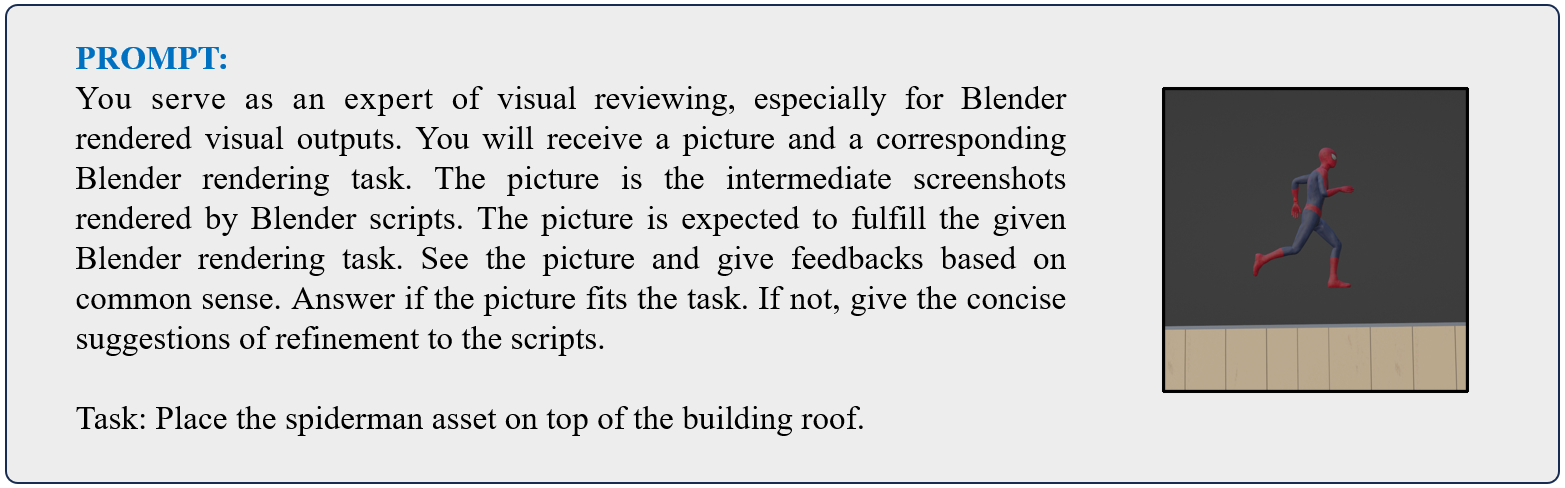}
  \centering
  \caption{\textbf{Prompt for Visual Reviewing.}  An example multimodal prompt for \texttt{VLM-Reviewer} to review the intermediate screenshot given the proposed Blender rendering task.}
  \label{fig:supp_prompt_reviewer}
\end{figure*}

We show examples of (1) the text prompt input to the \texttt{LLM-Director} agent to decompose a given video description into sub-processes (Fig.~\ref{fig:supp_prompt_director}); (2) the text prompt input to \texttt{LLM-Programmer} agent to generate a function for importing and scaling of a 3D asset (Fig.~\ref{fig:supp_prompt_programmer}); and (3) the multimodal prompt input to \texttt{VLM-Reviewer} agent to give feedback of the intermediate visual output given the proposed Blender rendering task (Fig.~\ref{fig:supp_prompt_reviewer}).


\section{Library Composing and Updating}
\begin{figure*}
  \centering
  \includegraphics[width=\linewidth]{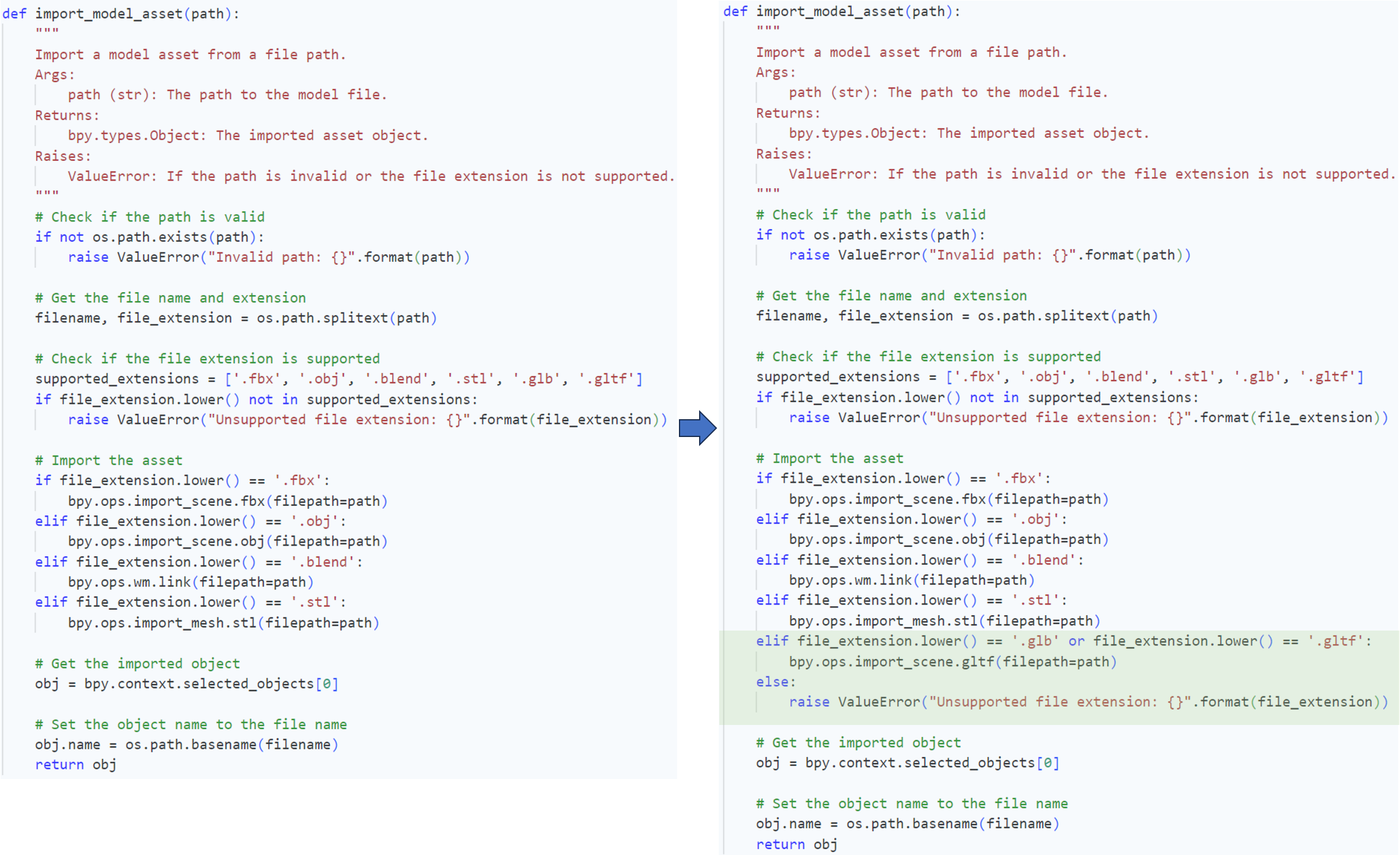}
  \centering
  \caption{\textbf{Library Updating of Asset Import Function.}  An example of library updating by \texttt{LLM-Programmer} to improve the former function to adapt more versatile file extensions (Green shaded area).}
  \label{fig:supp_scripts_input}
\end{figure*}

In Fig.~\ref{fig:supp_scripts_input}, we show an example of the library updating a function for importing 3D assets into the scene by \texttt{LLM-Programmer}. The initial function produces an error when importing an asset with the unknown ".glb" extension. The debugging errors are used update the function in the library and further improve the capability of the library.

\begin{figure*}
  \centering
  \includegraphics[width=0.8\linewidth]{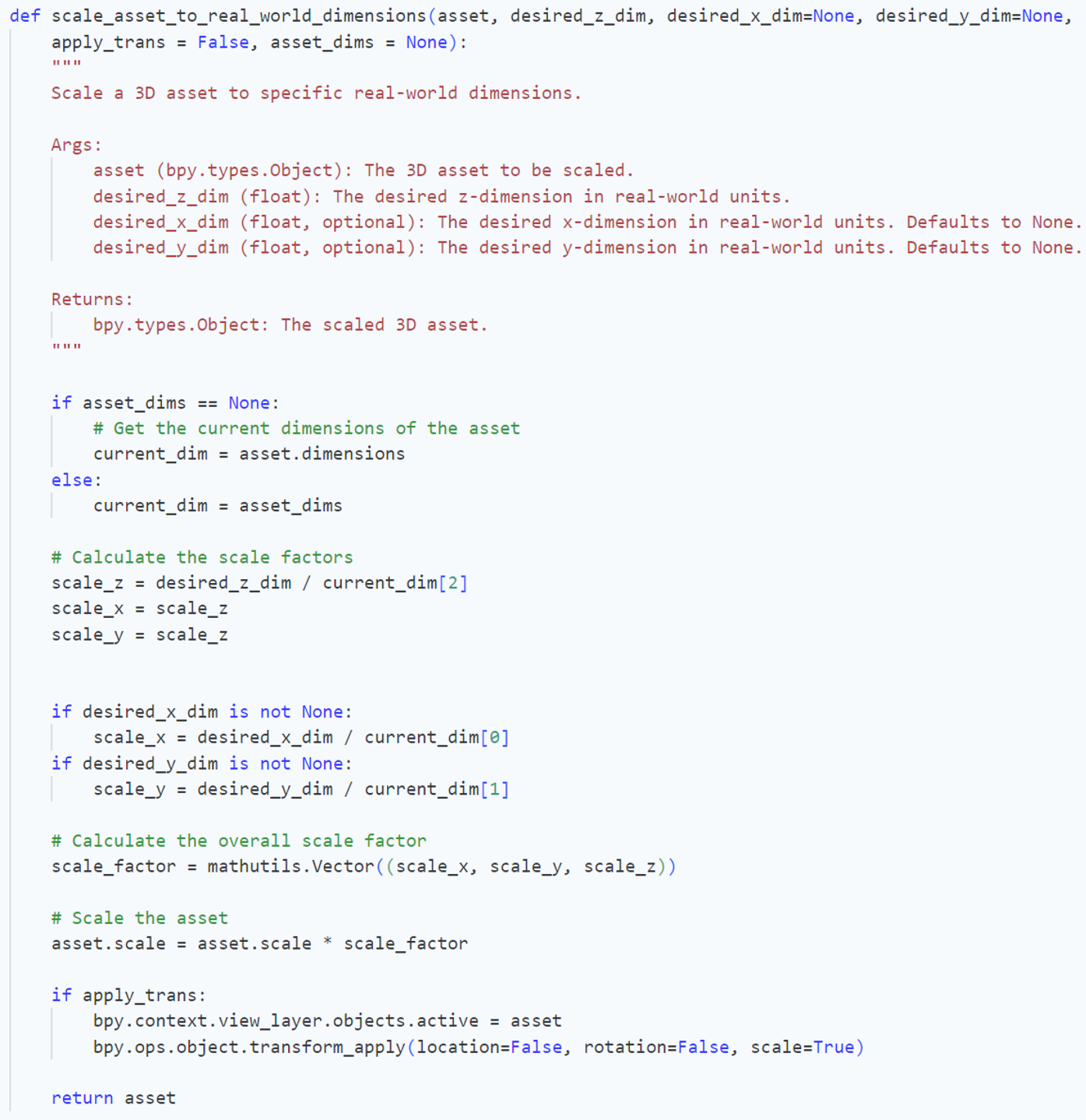}
  \centering
  \caption{\textbf{Function of Scaling.} An example function generated by \texttt{LLM-Programmer} to scale the given asset to its real-world size by a given reasonable Z dimension.}
  \label{fig:supp_scripts_scale}
\end{figure*}

In Fig.~\ref{fig:supp_scripts_scale}, we show the function for rotating 3D assets.

\begin{figure*}
  \centering
  \includegraphics[width=0.8\linewidth]{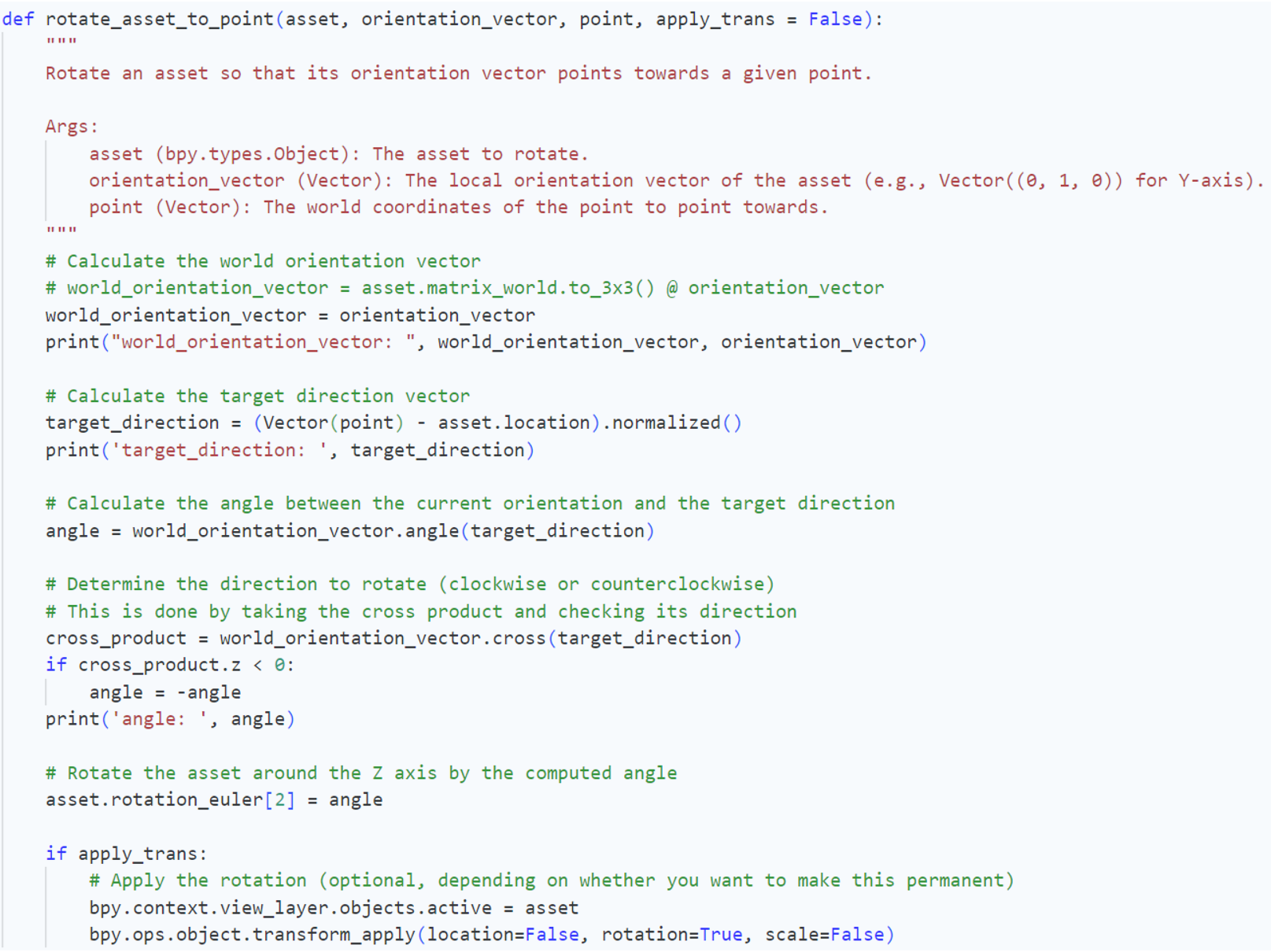}
  \centering
  \caption{\textbf{Function of Rotation.} An example function generated by \texttt{LLM-Programmer} to rotate a given asset in the XY-plane in order to face a given 3D point.}
  \label{fig:supp_scripts_rotate}
\end{figure*}

In Fig.~\ref{fig:supp_scripts_rotate}, we provide the function of scaling 3D assets to real world dimensions. 

\begin{figure*}
  \centering
  \includegraphics[width=0.8\linewidth]{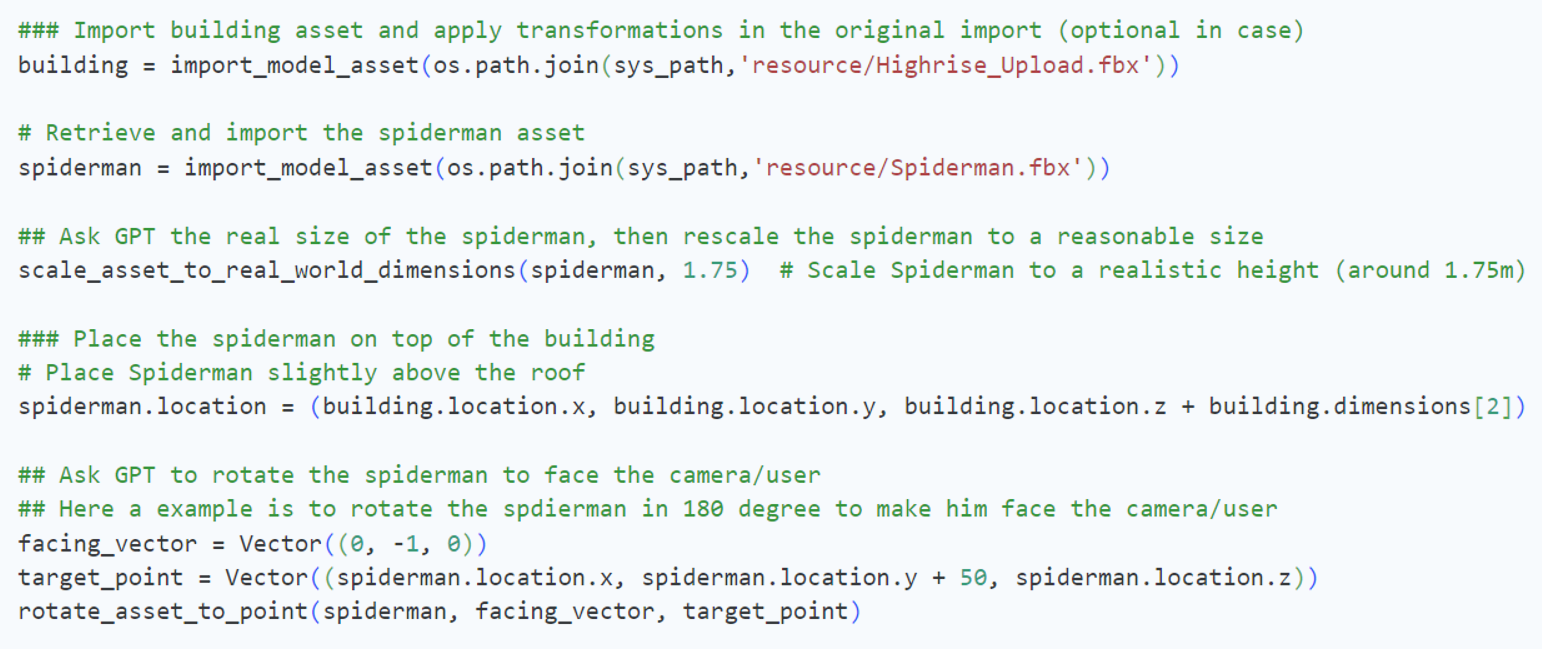}
  \centering
  \caption{\textbf{Code Snippet For Calling Library Functions.}  An example code snippet generated by \texttt{LLM-Programmer} calling the functions in library which are showed in Fig~\ref{fig:supp_scripts_input}, Fig.~\ref{fig:supp_scripts_scale}, and Fig.~\ref{fig:supp_scripts_rotate}. It fulfills the task: "generate a building and place Spiderman on top of the building roof. The Spiderman is facing the roof edge."}
  \label{fig:supp_scripts_spiderman_snippet}
\end{figure*}

In Fig.~\ref{fig:supp_scripts_spiderman_snippet}, we provide an example code snippet generated by \texttt{LLM-Programmer} calling the functions in the library which are showed in Fig~\ref{fig:supp_scripts_input}, Fig.~\ref{fig:supp_scripts_scale}, and Fig.~\ref{fig:supp_scripts_rotate}. It fulfills the task: "generate a building and place Spiderman on top of the building roof. The Spiderman is facing the roof edge."

\section{User Study}
\begin{figure*}[b]
  \centering
  \includegraphics[width=0.85\linewidth]{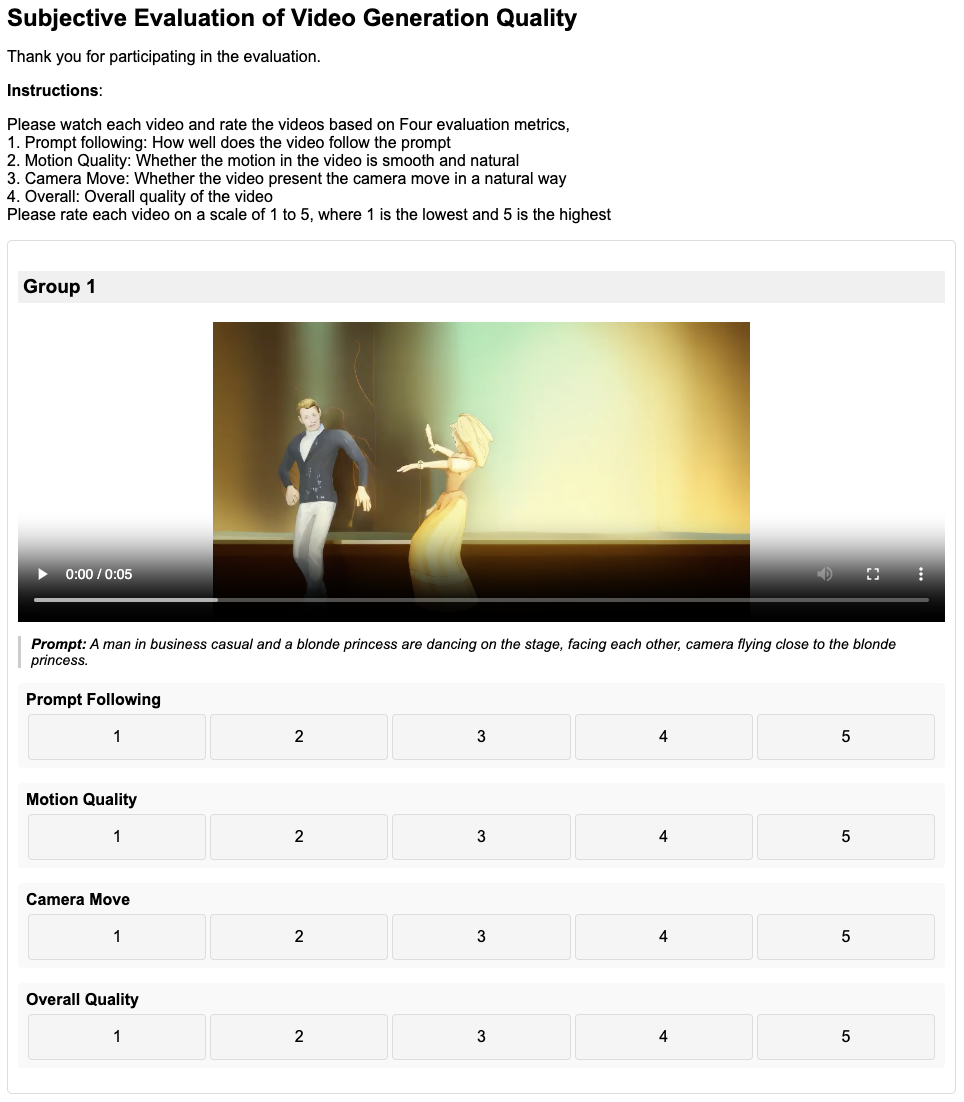}
  \centering
  \caption{\textbf{User Study Page.} An example user study page for evaluating video generation quality. Participants rate videos based on four metrics: Prompt Following, Motion Quality, Camera Move, and Overall Quality, on a scale of 1 to 5. Videos from our method and 7 other baselines are randomly placed. Black boxes hide logos in the upper right and lower right corners.
}
  \label{fig:supp_user_study}
\end{figure*}

In addition to the information about the user study provided in the main paper (Qualitative Comparisons Section), in Fig.~\ref{fig:supp_user_study} we show an example user study page containing videos generated by our method and seven baselines. The user will see all videos along with their descriptions and rate each video based on four criteria: prompt following, motion quality, camera move, and overall quality, on a scale of 1 to 5. Each user will evaluate a total of 10 pages similar to Fig.~\ref{fig:supp_user_study}. Note that we cover the upper right and lower right corners of all videos with same-sized black boxes to obscure baseline method watermarks.

\end{document}